\documentclass[letterpaper]{article}
\usepackage{aaai}
\usepackage{times}
\usepackage{helvet}
\usepackage{courier}

\usepackage{amsfonts}
\usepackage{bm}
\usepackage{amsmath}
\usepackage{amsthm}
\usepackage{array}
\usepackage{bbm}

\usepackage{longtable}
\usepackage{makecell}

\newtheorem{theorem}{Theorem}[section]

\newtheorem{corollary}[theorem]{Corollary}

\newtheorem{definition}[theorem]{Definition}

\begin{document}
%
\title{Generalization Analysis for Game-Theoretic Machine Learning }
\author{(Hide for Double-Blind Review)}

\maketitle
\begin{abstract}
\begin{quote}
For Internet applications like sponsored search, cautions need to be taken when using machine learning to optimize their mechanisms (e.g., auction) since self-interested agents in these applications may change their behaviors (and thus the data distribution) in response to the mechanisms. To tackle this problem, a framework called game-theoretic machine learning (GTML) was recently proposed, which first learns a Markov behavior model to characterize agents¡¯ behaviors, and then learns the optimal mechanism by simulating agents' behavior changes in response to the mechanism. While GTML has demonstrated practical success, its generalization analysis is challenging because the behavior data are non-i.i.d. and dependent on the mechanism. To address this challenge, first, we decompose the generalization error for GTML into the behavior learning error and the mechanism learning error; second, for the behavior learning error, we obtain novel non-asymptotic error bounds for both parametric and non-parametric behavior learning methods; third, for the mechanism learning error, we derive a uniform convergence bound based on a new concept called \emph{nested covering number} of the mechanism space and the generalization analysis techniques developed for mixing sequences. To the best of our knowledge, this is the first work on the generalization analysis of GTML, and we believe it has general implications to the theoretical analysis of other complicated machine learning problems.
\end{quote}
\end{abstract}

\section{Introduction}

Many Internet applications, such as sponsored search and crowdsourcing, can be regarded as dynamic systems that involve multi-party interactions. Specifically, \emph{users} arrive at the system at random with their particular needs; \emph{agents} provide products or services that could potentially satisfy users' needs; and the \emph{platform} employs a mechanism to match agents with users. Afterwards, users may give feedback to the platform about their satisfactions; the platform extracts revenue and may provide agents with some signals as their performance indicator. Since both the information reported by the agents and the mechanism will affect the payoff of the agents, self-interested agents may strategically adjust their behaviors (e.g., strategically report the information about their services or products) in response to the mechanism (or more accurately the signals they receive since the mechanism is invisible to them). Take sponsored search as an example. When a user submits a query to the search engine (the platform), the search engine runs an auction to determine a ranked list of ads based on the bid prices reported by the advertisers (the agents). If the user clicks on (gives feedback to) an ad, the search engine will charge the corresponding advertiser by a certain amount of money. After a few rounds of auctions, the search engine will provide the advertisers with some signals on the auction outcome, e.g., the average rank positions of their ads, the numbers of clicks, and the total payments. Based on such signals, the advertisers may adjust their bidding behaviors to be better off in the future.

It is clear that the mechanism plays a central role in the aforementioned dynamic system. It determines the satisfaction of the users, the payoffs of the agents, and the revenue of the platform. Therefore, how to optimize the mechanism becomes an important research topic. In recent years, a number of research works \cite{Lahaie07:Rev4RankingRules,radlinski2008optimizing,Zhu09:RevOptWithConst,Zhu09:OptRev,Mohri13:SPReserve,Di14:GTML,Fei14:ABPFull} have used machine learning to optimize the mechanism. These works could be categorized into three types.
\begin{itemize}
\item Some researchers assume that the agents are fully rational and investigate the Nash (or dominant-strategy) equilibrium of the mechanism. For example, \cite{Mohri13:SPReserve} proposes a machine learning framework to optimize the second-price auction in sponsored search in the single-slot setting. In this case, the dominant strategy for fully rational advertisers is to truthfully reveal their valuations through the bid prices and therefore their bidding behaviors have no dynamics.
\item Some researchers assume that the behaviors of the agents are i.i.d. and independent of the mechanism, and optimize the mechanisms based on historical behavior data. For example, \cite{Zhu09:RevOptWithConst} and \cite{Zhu09:OptRev} apply machine learning algorithms to optimize the first-price auction based on the advertisers' historical bidding data.
\item Some other researchers believe that the behaviors of the agents are neither fully rational nor i.i.d., instead, they are dependent on the mechanism through a data-driven Markov model. For example, \cite{HeCWL13:GTMLConf} and \cite{Fei14:ABPFull} assume that the agents' behavior change is Markovian, i.e., dependent on their historical behaviors and the received signals in previous time periods.
\end{itemize}
Please note that the assumption in the third type of works is more general, and can cover the other two types as its special cases. According to \cite{FudenburgeLearninginGames}, Nash (and also dominant-strategy) equilibrium in many games can be achieved by best-response behaviors, with which an agent determines the next action by maximizing his/her payoff based on the current action profile and mechanism. It is clear that the best-response behaviors are Markovian. Furthermore, it is also clear that the i.i.d. behaviors are special cases of Markov behaviors, where the Markov transition probability is reduced to a fixed distribution independent of the signals and the previous behaviors.

Based on the Markov assumption on agent behaviors, \cite{Di14:GTML} propose a new framework for mechanism optimization, called game-theoretic machine learning (GTML). The GTML framework involves a bi-level empirical risk minimization (ERM): it first learns a Markov model to characterize how agents change their behaviors, and then optimizes the mechanism by simulating agents' behavior changes in response to the mechanism based on the learned Markov model. The GTML framework has demonstrated promising empirical results, however, its generalization analysis is missing in the literature\footnote { In \cite{Fei14:ABPFull}, authors only studied the generalization ability for behavior learning. Furthermore, their definition for behavior learning error is different from ours, and cannot be applied to the generalization analysis for GTML.}. Actually this is a very challenging task because conventional machine learning assumes data is i.i.d. generated from an unknown but fixed distribution \cite{devroye1996probabilistic,vidyasagar2003learning}, while in GTML, agents¡¯ behavior data have time dependency and may dynamically change in response to the mechanism. As a result, conventional generalization analysis techniques could not be directly applied.

In this paper, we present a formal analysis on the generalization ability of GTML. Specifically, utilizing the stability property of the stationary distribution of Markov chain \cite{Mit05:mitrophanov2005sensitivity}, we decompose the generalization error for GTML into the behavior learning error and the mechanism learning error. The former relates to the process of learning a Markov behavior model from data, and the latter relates to the process of learning the optimal mechanism based on the learned behavior model. For the behavior learning error, we offer novel non-asymptotic error bounds for both parametric and non-parametric behavior learning methods: for parametric behavior learning method, we upper bound the behavior learning error by parameter learning error; for non-parametric behavior learning method, we derive a new upper bound for the gap between transition frequency and transition probability of a Markov chain. After that, we apply the Hoeffding inequality for Markov chains to both of the upper bounds, and obtain the error bound for both parametric and non-parametric behavior learning methods. For the mechanism learning error, we make use of a new concept called \emph{nested covering number} of the mechanism space. Specifically, we first partition the mechanism space into subspaces (i.e., a cover) according to the similarity between the stationary distributions of the data induced by mechanisms. In each subspace, the data distribution is similar and therefore one can substitute the data sample associated with each mechanism by a common sample without affecting the expected risk by much. Second, for each mechanism subspace, we derive a uniform convergence bound based on its covering number \cite{Ant09:anthony2009neural} by using the generalization analysis techniques developed for mixing sequences. In the end of this paper, we apply our generalization analysis of GTML to sponsored search, and give theoretical guarantee to GTML in this scenario.

To the best of our knowledge, this is the first work that performs formal generalization analysis on GTML, and we believe the methodologies we use have their general implications to the theoretical analysis of other complicated machine learning problems as well.

\section{GTML Framework}

In this section, we briefly introduce the game-theoretic machine learning (GTML) framework. For ease of reference, we summarize related notations in Table \ref{tbl_symbol}.

\subsection{Mechanisms in Internet Applications}

Internet applications such as sponsored search and crowdsourcing can be regarded as dynamic systems involving interactions between multiple parties, e.g., users, agents, and platform. For example, in sponsored search, the search engine (platform) ranks and shows ads to users and charges the advertisers (agents) if their ads are clicked by users, based on the relevance degrees of the ads and the bid prices reported by the advertisers. Similar multi-party relationship can also be found in crowdsourcing, where the platform corresponds to  Mechanical Turk, the agents corresponds to employers.  While we can assume the behaviors of the users to be i.i.d., the behaviors of the agents are usually not. This is because agents usually have clear utilities in their minds, and they may change behaviors in order to maximize their utilities given the understandings on the mechanism used by the platform. As a result, the agents' behaviors might be dependent on the mechanism.

Mathematically, we denote the space of mechanisms as $\mathcal{A}$, and assume it to be bounded with distance $d_{\mathcal{A}}$. We denote the space of user need/feedback, the space of agent behaviors, and the space of the signals, as $\mathcal{U}$, $\mathcal{B}$, and $\mathcal{H}$ respectively. We assume $\mathcal{B}$ and $\mathcal{H}$ are both finite, with size $|\mathcal{B}|$ and $|\mathcal{H}|$, since the behaviors and signals are usually discrete and bounded. For a mechanism $a\in \mathcal{A}$, at the $t$-th time period, agents' behavior profile is $b_t^a\in\mathcal{B}$, and a user $u_t\in\mathcal{U}$ arrives at the system. The platform matches the agents to the user and charges them, according to mechanism $a$. After that, the platform will provide some signals $h_t\in\mathcal{H}$ (e.g., the number of clicks on the ads ) to the agents as an indication of their performances. Since $h_t$ may be affected by agents' behavior profile $b_t^a$, mechanism $a$, and user data $u_t$, we denote $h_t = sig(a, b_{t}^a, u_t),$ where $sig: \mathcal{A}\times \mathcal{B}\times \mathcal{U}\to \mathcal{H}$ is a function generating the signals for agents. After observing $h_t$, agents will change their behavior to $b_{t+1}^a$ to be better off in the future.

\subsection{Markov Agent Behavior Model}

In order to describe how agents change their behaviors, the authors of \cite{HeCWL13:GTMLConf} and \cite{Fei14:ABPFull} proposed a Markov behavior model. The key assumption made by the Markov model is that any agent only has a limited memory, and his/her behavior change only depends on his/her previous behaviors and signals in a finite number of time periods. To ease the discussion and without loss of too much generality, they assume the behavior model to be first-order Markovian. Formally, given the signal $h_t$, the distribution of agents' next behavior profile can be written as follows,
\begin{small}
\begin{equation*}
\begin{aligned}
&P(b_{t+1}^a | b_t^a,...,b_1^a; u_t,...,u_1) \\
=& P(b_{t+1}^a  | b_t^a,...,b_1^a; h_t,...,h_1)\\
=&P(b^a_{t+1}|b_t^a,h_t) := M_{h_t}(b_t^a,b_{t+1}^a),
\end{aligned}
\end{equation*}
\end{small}
where $M_{h}$ is the transition probability matrix of the behavior profile, given the signals $h\in\mathcal{H}$.

As mentioned in the introduction, the Markov behavior model is very general and can cover other types of behavior models studied in the literature, such as the best-response behaviors and the i.i.d. behaviors.

\subsection{Bi-Level Empirical Risk Minimization}

In \cite{Di14:GTML}, bi-level empirical risk minimization (ERM) algorithm is proposed to solve the GTML problem. The first-level ERM corresponds to behavior learning, i.e., learning the Markov behavior model (the transition probability matrixes $M_{h}(\cdot,\cdot), h\in\mathcal{H}$) from training data containing signals and corresponding behavior changes. The second-level ERM corresponds to mechanism learning, i.e., learning the mechanism with the minimum empirical risk defined with both the behavior model learned at the first level and the training data containing users' needs/feedback.

For behavior learning, suppose we have $T_1$ samples of historical behaviors and signals $\{b_t, h_t\}_{t=1}^{T_1}$. The goal is to learn the transition matrix $M_{h}(\cdot,\cdot)$ from these data. In \cite{Di14:GTML} and \cite{Fei14:ABPFull}, both parametric and non-parametric approaches were adopted for behavior learning. With the parametric approach, one assumes the transition probability to take a certain mathematical form, e.g., $M_{h}(b,b')\propto \exp(-(b'-\langle w, (b,h,1)\rangle)^2)$, where $\langle\cdot,\cdot\rangle$ denotes the inner product of two vectors and parameter $w$ is learned by maximum likelihood estimation. With the non-parametric approach, one directly estimates each entry $M_{h}(b,b')$ by counting the frequencies of the event $(b_t=b,b_{t+1}=b')$ out of the event $(b_t=b)$ given signal $h$. No matter which approach is used, we denote the learned behavior model as $\hat{M}_{T_1}$ for ease of reference.

For mechanism learning, suppose we have $T_2$ samples of user data $\{u_t\}_{t = 1}^{T_2}$ and a Markov behavior model $\hat{M}_{T_1}$, learned as above. The goal is to learn an optimal mechanism to minimize the empirical risk (e.g., minus empirical revenue/social welfare) on the user data, denoted as $L(a,b,u)$ where $L:\mathcal{A}\times \mathcal{B} \times \mathcal{U} \to [-K,0]$. For this purpose, for arbitrary mechanism $a\in\mathcal{A}$, one generates $T_2$ samples of behavior data $\{b_t^a\}_{t=1}^{T_2}$ in a sequential manner using the Markov model $\hat{M}_{T_1}$ and $T_2$ samples of user data. With the $T_2$ samples of behavior data and user data, the empirical risk of mechanism $a$ can be computed. To improve the computational efficiency of mechanism learning, in \cite{Di14:GTML}, the authors introduce a technique called \emph{$\delta$-sample sharing}. Specifically, given $\delta>0$, in the optimization process, if the distance between a new mechanism $a$ and another mechanism $a'$ whose behavior data is already generated is smaller than $\delta$ (i.e., $d_{\mathcal{A}}(a,a')\leq \delta$), then one will not generate behavior data for $a$ any more, but instead reuse the behavior data previously generated for mechanism $a'$. Therefore, we denote the sample for mechanism $a$ as $\{b_t^{s(a, \delta)}\}_{t = 1}^{T_2}$, where $s(a, \delta)$ is equal to $a$ itself or another mechanism satisfying $d_{\mathcal{A}}(a, s(a, \delta)) \leq \delta$. Consequently, the empirical risk of mechanism $a$ is defined as below,
\begin{small}
\begin{equation*}
\mathcal{R}_{T_2}(a,\hat{M}_{T_1},\delta)= \frac{1}{T_2} \sum_{t=1}^{T_2} L(a, b_t^{s(a, \delta)}, u_t).
\end{equation*}
\end{small}
 By minimizing $\mathcal{R}_{T_2}(a,\hat{M}_{T_1},\delta)$, one can obtain an empirically optimal mechanism:
\begin{small}
\begin{equation*}
\label{eqn_a_hat}
\hat{a}_{T_2} =\arg \min_{a \in \mathcal{A}}\mathcal{R}_{T_2}(a,\hat{M}_{T_1},\delta).
\end{equation*}
\end{small}
While GTML and the bi-level ERM algorithm have demonstrated their practical success \cite{HeCWL13:GTMLConf}, their theoretical properties are not yet well understood. In particular, given that GTML is more complicated than conventional machine learning (in GTML the behavior data are time-dependent and mechanism-dependent), conventional generalization analysis techniques cannot be directly applied and new methodologies need to be proposed.

\section{Generalization Analysis for GTML}

In this section, we first give a formal definition to the generalization error of the bi-level ERM algorithm for GTML, and then discuss how to derive a meaningful upper bound for this generalization error. Finally, we apply our generalization analysis of GTML to sponsored search, and show the GTML in this scenario has good generalization ability.

According to \cite{Fei14:ABPFull}, for a behavior model $M$ (such as the true Markov behavior model $M^*$ and the model $\hat{M}_{T_1}$ obtained by the behavior learning algorithm), under some mild conditions (e.g., is irreducible and aperiodic), the process $(b_t^a,u_t)$ is a uniformly ergodic Markov chain for arbitrary mechanism $a$.  Then given mechanism $a$ and behavior model $M$, there exists a stationary distribution for $(b_t^a, u_t)$, which we denote as $\pi(a,M)$. For simplicity, we assume the process $\{(b_t^a,u_t):t\geq 1\}$ is stationary \footnote{Our results can similarly holds without this assumption.}. We define the risk for each mechanism $a \in \mathcal{A}$ as the expected loss with respect to the stationary distribution of this mechanism under the true behavior model $M^*$, i.e.,
\begin{small}
\begin{equation*}
\mathcal{R}(a,M^*)= E_{\pi(a,M^*)}L(a,b^a,u).
\end{equation*}
\end{small}
The optimal mechanism minimizing this risk is denoted as $a^*$, i.e.,
\begin{small}
\begin{equation*}
a^*=\arg\min_{a\in\mathcal{A}}\mathcal{R}(a,M^*).
\end{equation*}
\end{small}
We consider the gap between the risk of the mechanism $\hat{a}_{T_2}$ learned by the bi-level ERM algorithm and the  risk of the optimal mechanism $a^*$, i.e., $\mathcal{R}(\hat{a}_{T_2}, M^*) - \mathcal{R}(a^*, M^*)$.  We call this gap the generalization error for the bi-level ERM algorithm, or simply the generalization error for GTML.

To ease the analysis, we utilize the stability property of the stationary distribution of uniformly ergodic Markov Chain and decompose the generalization error for GTML into two parts, as shown in the following Theorem. Due to space restrictions, we leave all proofs in this paper to supplemental materials.
\begin{theorem}
\label{BehaviorUnknownTheorem}
The generalization error of the bi-level ERM algorithm for GTML $\mathcal{R}(\hat{a}_{T_2}, M^*) - \mathcal{R}(a^*, M^*)$ can be bounded as:
\begin{small}
\begin{equation}
\label{eqn_error_decompose}
\begin{aligned}
\mathcal{R}(\hat{a}_{T_2},M^*) - &\mathcal{R}(a^*,M^*)
\leq 2K C(M^*)||M^*- \hat{M}_{T_1}||_{\infty}\\
+ &2 \sup_{a \in \mathcal{A}} |\mathcal{R}(a,\hat{M}_{T_1}) - \mathcal{R}_{T_2}(a,\hat{M}_{T_1}, \delta)|,
\end{aligned}
\end{equation}
\end{small}
where $K$ is an upper bound for loss $L$, and $C(M^*)$ is a non-negative constant depending on $M^*$.
\end{theorem}
For ease of reference, we call the first term $||M^*- \hat{M}_{T_1}||_{\infty}$ in the right-hand side of inequality (\ref{eqn_error_decompose})  \emph{behavior learning error} and the second term $\sup_{a \in \mathcal{A}} |\mathcal{R}(a,\hat{M}_{T_1}) -\mathcal{R}_{T_2}(a,\hat{M}_{T_1}, \delta)|$  \emph{mechanism learning error}. We will derive upper bounds for both errors in the following subsections.

\subsection{ Error Bound for Behavior Learning}
\label{Error Bound for Behavior Learning}

In this subsection we derive  error bounds for both parametric and non-parametric behavior learning methods. Since the behavior space and signal space are both finite, it is shown in \cite{Fei14:ABPFull} that $\{b_{t+1}, b_t, h_t\}$ forms a time-homogeneous Markov chain. Furthermore, under regular conditions, the Markov chain is uniformly ergodic, i.e., there exists $N_0$, such that the elements in the $N_0$-step transition probability matrix of $\{b_{t+1}, b_t, h_t\}$ are all positive. For ease of reference, we denote the minimum element in this matrix as $\delta_0$. Since the mechanism $a_0$ is fixed in the process of behavior learning , we omit all the super scripts $a_0$ in $b_t^{a_0}$ if without confusion. Please note that, in \cite{Fei14:ABPFull}, although authors studied the generalization analysis for behavior learning, their definition on behavior learning error is different from ours and cannot be applied in the generalization analysis for GTML. To be specific, they measure the behavior learning error by the expected behavior prediction loss of the learned behavior model with respect to the stationary distribution under the true behavior model, while we measure behavior learning error in a stricter way by the infinity distance between the learned model and the true model.

\subsubsection{Parametric Behavior Learning}

With the parametric approach \cite{Di14:GTML}, the transition probability is proportional to a truncated Gaussian function, i.e., $M_h(b',b)\propto exp(-(b'-\langle\omega,(b,h,1)\rangle)^2)$ where $\omega$ is bounded . The parameter is obtained by maximizing the likelihood. We first bound the behavior learning error by the gap between the learned parameter and the parameter in the true model, utilizing the property of maximum likelihood method; then we apply the Hoeffding inequality for uniformly ergodic Markov chain \cite{Glynn02:hoeffding} and finally obtain the error bound for parametric behavior learning method as shown in the following theorem.
\begin{theorem}
\label{BehaviroErrThmParametric}
For any $\epsilon >0$, we have, for $T_1 > (2C_1N_0)/(|\mathcal{B}|^2|\mathcal{H}|\delta_0\epsilon)$,
\begin{small}
\begin{equation*}
\begin{aligned}
&P(||\hat{M}_{T_1}- M^*||_{\infty} \geq \epsilon)\\
\leq &2\exp\big(-\frac{\big(T_1\epsilon|\mathcal{B}|^2|\mathcal{H}|\delta_0-2C_1N_0\big)^2}{2T_1N_0^2C_1^2}\big)
=O(\exp(-T_1))
\end{aligned}
\end{equation*}
\end{small}
where $\delta_0,N_0,C_1$ are positive constants.
\end{theorem}
\subsubsection{Non-parametric Behavior Learning}

In the non-parametric behavior learning, we estimate the transition probability $M_{H_j}(B_i,B_k)$ by the conditional frequency of the event $b_{t+1} = B_k$ given that $b_t = B_i$ and $h_t = H_j$, i.e.,
$\hat{M}_{H_j}(B_i,B_k)=\frac{\sum_{t = 1}^{T_1}\mathbbm{1}_{\{b_{t+1} = B_k, b_t = B_i, h_t = H_j\}}}{\sum_{t = 1}^{T_1}\mathbbm{1}_{\{b_t = B_i, h_t = H_j\}}}.$
The difficulty in analyzing the error of the above estimation comes from the sum of random variables in the denominator of the conditional frequency. To tackle the challenge, we first derive an upper bound for the gap between conditional transition frequency and conditional transition probability, which does not involve such a sum of random variables, then apply the Hoeffding inequality for uniformly ergodic Markov chain \cite{Glynn02:hoeffding} to this upper bound. In this way, we manage to obtain a behavior learning error bound, as shown in the following theorem.

\begin{theorem}
For any $\epsilon >0$, we have for $T_1>\big(2N_0(|\mathcal{B}|+1)\big)/\big(|\mathcal{B}||\mathcal{H}|\delta_0 C_2\epsilon\big)$,
\label{BehaviorErrThm}
\begin{small}
\begin{equation*}
\begin{aligned}
&P(||\hat{M}_{T_1}- M^*||_{\infty} \geq \epsilon) \\
\leq &2|\mathcal{H}||\mathcal{B}|^2(|\mathcal{B}|+1) \exp\big(- \frac{\big(C_2T_1\delta_0|\mathcal{B}||\mathcal{H}|\epsilon - 2N_0(|\mathcal{B}|+1)\big)^2}{2T_1N_0^2(|\mathcal{B}|+1)^2}\big)\\
=&O(\exp(-T_1))
\end{aligned}
\end{equation*}
\end{small}
where $\delta_0,N_0,C_2$ are positive constants.
\end{theorem}
\subsection{Error Bound for Mechanism Learning}
\label{MainGeneralization}

In this section, we bound the mechanism learning error by using a new concept called \emph{nested covering number} for the mechanism space. We first give its  definition, and  then prove a uniform convergence bound for mechanism learning on its basis.

\subsubsection{Nested Covering Number of Mechanism Space}

The nested cover contains two layers of covers: the first-layer cover is defined for the entire mechanism space based on the distance between stationary distributions induced by the mechanisms. The second-layer cover is defined for each partition (subspace) obtained in the first layer based on the distance between the losses of the mechanisms projected onto finite common data samples.

First, we construct the first-layer cover for the mechanism space $\mathcal{A}$.
In mechanism learning, the learned Markov behavior model $\hat{M}_{T_1}$ is used to generate the behavior data for different mechanisms. For simplicity, we denote the stationary distribution of the generated data as $\pi(a,\hat{M}_{T_1})$ (or $\pi_a$ for simplification) and the set of stationary distributions for $\mathcal{A}$ as $\pi(\mathcal{A})$. We define the (induced) total variance distance on $\mathcal{A}$ as the total variance distance on $\pi(\mathcal{A})$, i.e., for $\forall a, a' \in \mathcal{A}$, $d_{TV}(a,a')=d_{TV}(\pi_a,\pi_{a'})$. For $\forall \epsilon >0$, the smallest $\epsilon$-cover of $\mathcal{A}$ w.r.t. the total variance distance is $cover_{TV}^\epsilon = \{g_1^\epsilon, \cdots, g_i^\epsilon,\cdots\}$, where $g_i^\epsilon\in\mathcal{A}$. That is, $\mathcal{A}\subseteq \cup_i B(g_i^\epsilon,\epsilon)$, where $B(g_i^\epsilon,\epsilon)$ is the $\epsilon$-balls of $g_i^\epsilon$ with respect to the (induced) total variance distance. We define \emph{the first-layer covering number} as the cardinality of $cover_{TV}^\epsilon$, denoted as $ \mathcal{N}_{TV}(\epsilon, \mathcal{A})$. Based on $cover_{TV}^\epsilon$, we can obtain a partition for $\mathcal{A}$, denoted as $\{\mathcal{A}_i^\epsilon\}$, where $\mathcal{A}_i^\epsilon$ is an $\epsilon$-partition of $\mathcal{A}$. When the mapping from mechanism to its stationary distribution is $\alpha$ uniformly Lipschitz continuous, then $\mathcal{N}_{TV}(\epsilon, \mathcal{A}) < \infty$. Because for $\forall \delta>0$,  $a$ and $s(a,\delta)$ belong to the same $\delta \alpha$-partition of $\mathcal{A}$. So, considering $\mathcal{A}$ is bounded, we have $\mathcal{N}_{TV}(\delta \alpha, \mathcal{A}) \leq \mathcal{N}_{d_{\mathcal{A}}}(\delta, \mathcal{A}) < \infty$.

Second, we consider the loss functions for each mechanism subspace $L\circ \mathcal{A}_i^{\epsilon} := \{L \circ a: \mathcal{B}\times \mathcal{U} \rightarrow [-K,0]| L\circ a(b_t^a,u_t) = L(a,b_t^a,u_t), a\in \mathcal{A}_i^{\epsilon}\}$, and define its covering number w.r.t.  the $T_2$ common samples  $\{X_{i,t}^\epsilon\}_{t=1}^{T_2}$, where $X_{i,t}^\epsilon=\{ b^{g_i^{\epsilon}}_t,u_t \}$ and $\{b_t^{g_i^{\epsilon}}\}_{t=1}^{T_2}$ are generated by mechanism $g_i^\epsilon$. Again, we define the second-layer cover as the smallest $\epsilon'$-cover of $L\circ \mathcal{A}_i^{\epsilon}|_{\{X_{i,t}^\epsilon\}_{t=1}^{T_2}}$ under the $l_1$ distance, i.e., $cover_1^{\epsilon'}(L\circ \mathcal{A}_i^{\epsilon}|_{\{X_{i,t}^\epsilon\}_{t=1}^{T_2}})$, and define \emph{the second-layer covering number} $\mathcal{N}_1(\epsilon',L\circ\mathcal{A}_i^{\epsilon},T_2)$ as its maximum cardinality with respect to the sample $\{X_{i,t}^\epsilon\}_{t=1}^{T_2}$.

In summary, the nested covering numbers for a mechanism space are defined as follows:
\begin{definition}
Suppose $\mathcal{A}$ is a mechanism space, we define its nested covering numbers as $\big\{\mathcal{N}_{TV}(\epsilon, \mathcal{A}), \{\mathcal{N}_1(\epsilon',L \circ \mathcal{A}_i^{\epsilon},T_2)\}\big\}$.
\end{definition}

\subsubsection{Uniform Convergence Bound for Mechanism Learning}
\label{uniform convergence part}

In this subsection, we derive a uniform convergence bound for the ERM algorithm for mechanism learning. We first relate the uniform convergence bound for the entire mechanism space to that for the subspaces constructed according to the first-layer cover. Then considering that uniformly ergodic Markov chains are $\beta$-mixing \cite{doob90:stochastic}, we make use of the \emph{independent block technique} for mixing sequences \cite{Yu94:yu1994rates} to transform the original problem based on dependent samples to that based on independent blocks. Finally, we apply the symmetrization technique and Hoeffding inequality to obtain the desired bound.
\begin{theorem}
\label{uniform convergence bound}
Suppose that the mapping from $\mathcal{A}$ to $\pi_{\mathcal{A}}$ is $\alpha$ uniformly Lipschitz continuous, and the $\beta$-mixing rate of Markov chain $\{(b_t^a,u_t):t\geq 1\}$ (denoted as $\beta(a,m)$) is algebraical(i.e., $\beta(a,m) \leq \beta_0 m^{- \gamma}$, where $\beta_0, \gamma \geq 0, m \in \mathbb{Z}$.). For any $\epsilon>0$,  we have
\begin{small}
\begin{equation*}
\label{eqn_uniform_converge_example}
\begin{aligned}
&P(\sup_{a \in \mathcal{A}}|\mathcal{R}_{T_2}(a,\hat{M}_{T_1},\delta) - \mathcal{R}(a,\hat{M}_{T_1})| \geq \epsilon)\\
\leq &\mathcal{N}_{d_{\mathcal{A}}}(\delta, \mathcal{A})\max_{1\leq i \leq \mathcal{N}_{d_{\mathcal{A}}}(\delta, \mathcal{A})}\big(16 \mathcal{N}_1((\epsilon-K\alpha\delta)/16, L \circ \mathcal{A}_i^{\delta},T_2)\\
\end{aligned}
\end{equation*}
\end{small}
\begin{small}
\begin{equation*}
\begin{aligned}
&\exp\big(- \frac{(\epsilon-\delta \alpha K)^2}{128K^2}\lceil\frac{T_2^{\frac{s}{1+s}}}{2}\rceil\big)+ \beta_0\lceil T_2^{\frac{s - \gamma}{1+s}}\rceil\big)\\
=O\bigg(\mathcal{N}_{d_{\mathcal{A}}}(\delta, \mathcal{A})&\big(\mathcal{N}_1(\epsilon, L \circ \mathcal{A},T_2)e^{-T_2^{\frac{s}{1+s}}}+T_2^{\frac{s - \gamma}{1+s}}\big)\bigg),
\end{aligned}
\end{equation*}
\end{small}
where $\lceil\rceil$ denotes ceiling function, $\delta\in (0,\epsilon/(K\alpha))$ and $s \in (0,\gamma)$.
\end{theorem}
\emph{Remark 1}: For space restrictions, we only present the bound with specific mixing rate, which is simpler and easier to understand. Without the assumption on the mixing rate, we can also obtain a similar bound, which can be found in Theorem C.1 in the supplemental materials.

\emph{Remark 2}: Although we have to leave the proofs to the supplementary materials due to space restrictions, we would like to point out one particular discovery from our proofs. While the $\delta$-sample sharing technique was originally proposed to improve efficiency, according to our proof it plays an important role in generalization ability . Then a question is whether this technique is necessary for generalization ability. Our answer is yes if $\mathcal{A}$ is infinite. Let us consider a special case in which $\mathcal{U} = \{u\}$ and $\pi_a\equiv\pi, \forall a\in\mathcal{A}$, i.e., the behavior model does not rely on the signals. If $\delta$-sample sharing is not used, for finite $T$,
\begin{small}
\begin{equation*}
\begin{aligned}
&P(\sup_{a \in \mathcal{A}} |\frac{1}{T}\sum_{t=1}^T L(a, b_t^a, u_t) - E L(a, b_t^a, u_t)| \geq \epsilon)\\
= &1 - \prod_{a\in\mathcal{A}}\big(1 - P(|\frac{1}{T}\sum_{t=1}^T L(a, b_t^a, u_t) - E L(a, b_t^a, u_t)|)\big) = 1.
\end{aligned}
\end{equation*}
\end{small}
This implies that mechanism learning without $\delta$-sample sharing does not have generalization ability.

\emph{Remark 3}: An assumption made in our analysis  is that the map from $\mathcal{A}$ to $\pi_{\mathcal{A}}$ is uniformly Lipschitz continuous. However, sometimes this assumption might not hold. In this case, we propose a modification to the original $\delta$-sample sharing technique. The modification comes from the observation that the first-layer cover is constructed based on the total variance distance between stationary distributions of mechanisms. Therefore, in order to ensure a meaningful cover, we could let two mechanisms share the same data sample if the estimates of their induced stationary distributions (instead of their parameters) are similar. Please refer to the supplementary materials for details of this modification and a proof showing how it can bypass the discontinuity challenge. Note that the modified $\delta$-sample sharing technique no longer has efficiency advantage since it involves the generation of behavior data for every mechanism examined during the training process, however, it ensures the generalization ability of the mechanism learning algorithm, which is desirable from the theoretic perspective.

\subsection{The Total Error Bound}

By combining Theorem \ref{BehaviorUnknownTheorem}, Theorem \ref{BehaviorErrThm}, Theorem \ref{BehaviroErrThmParametric} and Theorem \ref{uniform convergence bound}, we obtain the total error bound for GTML as shown in the following theorem.
\begin{theorem}
\label{totalerrorbound}
With the same assumptions in Theorem \ref{uniform convergence bound}, for bi-level ERM algorithm in GTML, for any $\epsilon>0$, we have the following generalization error bound \footnote{Please refer to Theorem C.3 for the total error bound without the assumption on the mixing rate.}:
\begin{small}
\begin{equation*}
\begin{aligned}
&P(\mathcal{R}(\hat{a}_{T_2},M^*) - \mathcal{R}(a^*,M^*) \geq \epsilon)\\
\leq&O(e^{-T_1})+O\bigg(\mathcal{N}_{d_{\mathcal{A}}}(\delta, \mathcal{A})\big(\mathcal{N}_1(\epsilon,L \circ \mathcal{A},T_2)e^{-T_2^{\frac{s}{1+s}}}+T_2^{\frac{s - \gamma}{1+s}}\big)\bigg),
\end{aligned}
\end{equation*}
\end{small}
where $ s\in (0,\gamma)$, and $\delta\in (0,\epsilon/K\alpha)$.
\end{theorem}
From the above theorem, we have following observations: 1) The error bound will converge to zero when the scales of agent behavior data $T_1$ and user data $T_2$ approach infinity. 2) The convergence rate w.r.t. $T_1$ is faster than that w.r.t. $T_2$, indicating that one needs more user data than agent behavior data for training. 3) The mechanism space impacts the generalization ability through both its first layer covering number (which is finite) and second layer covering number.

\subsection{Application to Sponsored Search Auctions}

In this section, we apply our generalization analysis for GTML to sponsored search auctions. In sponsored search, GSP auctions with a query-dependent reserve price are widely used \cite{Edelman05:GSP,Eas10:easley2010networks,Mohri13:SPReserve}.

When a reserve price $r\in \mathcal{R}^+$ is used, the GSP auction runs in the following manners. First, the search engine ranks the ads according to their bid prices (here we follow the common practice to absorb the click-through rate of an ad into its bid price to ease the notations), and will show to the users those ads whose bid prices are higher than the reserve price. If the ad on the $i$-th position (denoted as $ad_i$) is clicked by a user, the search engine will charge the corresponding advertiser by the maximum of the bid price of $ad_{i+1}$ and the reserve price $r$. For sake of simplicity and without loss of generality, we will only consider two ad slots. Let the binary vector $c=\{c_1, c_2\}$ indicate whether $ad_1$ and $ad_2$ are clicked by users.  Then the user data include two components, i.e., $u=(q, c)$, where $q\in Q$ is the query issued by the user and $c$ records user's click feedback.  Denote the bid profile of the shown ads as $(b^{(1),q}$, $b^{(2),q})$ (for simplicity we sometimes omit $q$ in the notation).  We consider a query-dependent reserve price, i.e., the auction family is $\mathcal{A} = \{a: Q \rightarrow \mathbb{R}^+\}$. For a mechanism $a$, the revenue of the search engine can be represented as:
\begin{small}
\begin{equation*}
\label{revenueofreserve}
\begin{aligned}
&Rev(a,b,u) = a(q)c_1\mathbbm{1}_{\{b^{(2)}\leq a(q) \leq b^{(1)}\}} \\
+ &(b^{(2)}c_1+ a(q)c_2)\mathbbm{1}_{\{b^{(3)}\leq a(q)\leq b^{(2)}\}}
+ (b^{(2)}c_1+ b^{(3)}c_2)\mathbbm{1}_{\{a(q)\leq b^{(3)}\}},
\end{aligned}
\end{equation*}
\end{small}
and the loss is $L(a,b,u)=-Rev(a,b,u)$.

Since the first layer covering number is always finite and independent of the user data size (i.e., $T_2$), we just need to bound the second-layer covering number for GSP auctions space with reserve price, which is shown as below.
\begin{theorem}
\label{finalcoveringnumbertheorem}
For GSP auctions with reserve price, the second layer covering number can be bounded by the pseudo-dimension (P-dim) of the reserve price function class. To be specific, we have:
\begin{small}
\begin{equation*}
\mathcal{N}_1(\epsilon', L\circ\mathcal{A}_i^{\epsilon},T_2) \leq (eT_2K/\epsilon')^{16|\mathcal{B}|\text{P-dim}(\mathcal{A})},\forall T_2 > 4|\mathcal{B}|\text{P-dim}(\mathcal{A}).
\end{equation*}
\end{small}
\end{theorem}
Combine Theorem \ref{totalerrorbound} and Theorem \ref{finalcoveringnumbertheorem}, we get a total error bound for GTML applied to GSP auctions with reserve price in the following theorem, which first gives generalization guarantees for GTML in sponsored search.
\begin{corollary}
With the same assumptions in Theorem \ref{uniform convergence bound}, for any $\epsilon>0$, for GTML applied to GSP auctions with reserve price, we have the following generalization error bound:
\begin{small}
\begin{equation*}
\begin{aligned}
&P(\mathcal{R}(\hat{a}_{T_2},M^*) - \mathcal{R}(a^*,M^*) \geq \epsilon)\\
\leq &O(e^{-T_1})+O\bigg(\mathcal{N}_{d_{\mathcal{A}}}(\delta, \mathcal{A})\big(T_2^{16|\mathcal{B}|\text{P-dim}(\mathcal{A})}e^{-T_2^{\frac{s}{1+s}}}+T_2^{\frac{s - \gamma}{1+s}}\big)\bigg),
\end{aligned}
\end{equation*}
\end{small}
where $ s\in (0,\gamma)$, and $\delta\in (0,\epsilon/K\alpha)$.
\end{corollary}

\section{Conclusion and Future Work}

In this paper, we have given a formal generalization analysis to the game-theoretic machine learning (GTML) framework, which involves a bi-level ERM learning process (i.e., mechanism learning and behavior learning). The challenges of generalization analysis for GTML lies in the dependency between the behavior data and the mechanism. To tackle the challenge, we first bound the error of behavior learning by leveraging the Hoeffding inequality for Markov Chains, and then introduce a new notion called \emph{nested covering number} and bound the errors of mechanism learning on its basis. Our theoretical analysis not only enriches the understanding on machine learning algorithms in complicated dynamic systems with multi-party interactions, but also provides some practical algorithmic guidance to mechanism design for these systems. As for future work, we would also like to extend the idea of $\delta$-sample sharing and apply it to improve the mechanisms in other real-world applications, such as mobile apps and social networks.

\begin{table}[htpb]
\footnotesize
\begin{tabular}{|c|p{140pt}|}
\hline
Notation & \makecell[c]{Meaning} \\ \hline
$\mathcal{U},\mathcal{B},\mathcal{H}$ &Spaces of user need/feedback, agent behaviors, and signals\\ \hline
$\mathcal{A}, d_{\mathcal{A}}$ &mechanism space and the distance on it\\ \hline
$u_t, b_t^a,h_t$ &at the $t$-th time period, under mechanism $a$, the users¡¯ need/feedback, agents' behavior, and the signal\\ \hline
$M_h(\cdot,\cdot)$ & transition probability matrix of agents¡¯ behavior under signal $h$\\ \hline
$\hat{M}_{T_1}, M^*$ &the learned behavior model and the true behavior model \\ \hline
$\hat{a}_{T_2}, a^*$ &the learned mechanism and the optimal mechanism \\ \hline
$L(a,b,u)$ &loss function \\ \hline
$s(a,\delta)$ &a mechanism that is equal to $a$ or another mechanism satisfying $d_{\mathcal{A}}(a,s(a,\delta))\leq \delta$ \\ \hline
$\pi(a,M)$ &stationary distribution of the process $(b_t^a,u_t)$ with behavior model $M$ \\ \hline
$\mathcal{R}_{T_2}(a,\hat{M}_{T_1},\delta)$ &empirical risk of mechanism $a$ with behavior model $\hat{M}_{T_1}$ by $\delta$-sample sharing technique \\ \hline
$\mathcal{R}(a,M^*)$ &expected risk of mechanism $a$ with the true behavior model $M^*$ \\ \hline
$d_{TV}(a,a')$ &(induced) total variance distance on mechanism space $\mathcal{A}$ \\ \hline
$\mathcal{N}_{d_{\mathcal{A}}}(\delta,\mathcal{A}), \mathcal{N}_{TV}(\epsilon, \mathcal{A})$ &covering number of mechanism space $\mathcal{A}$ under distance $d_{\mathcal{A}}$ and (induced) total variance distance $d_{TV}$ \\ \hline
$\mathcal{A}_i^{\epsilon}$ &the $\epsilon$-partition of mechanism space $\mathcal{A}$ according to its first layer cover\\ \hline
$L\circ \mathcal{A}_i^{\epsilon}$ &the loss function class in each partition\\ \hline
$\mathcal{N}_1(\epsilon,L\circ\mathcal{A}_i^{\epsilon},T_2)$ &covering number for the function class $L\circ \mathcal{A}_i^{\epsilon}$ under $l_1$ distance\\ \hline
$\beta(a,m)$ &beta mixing rate of Markov chain $\{(b_t^a,u_t):t\geq 1\}$ \\ \hline
\end{tabular}
\caption{Notations}
\label{tbl_symbol}
\end{table}

\newpage

\bibliographystyle{aaai}
\bibliography{ref}

\clearpage

\end{document}